\documentclass[12pt]{article}

\usepackage[english]{babel}
\usepackage[utf8x]{inputenc}
\usepackage[T1]{fontenc}
\usepackage{soul}
\usepackage[a4paper]{geometry}

\usepackage{graphicx}
\usepackage[colorinlistoftodos]{todonotes}
\usepackage[colorlinks=true, allcolors=blue]{hyperref}
\usepackage{multirow}
\usepackage{color}
\usepackage[normalem]{ulem}
\usepackage{authblk}
\usepackage{booktabs}

\title{Integrated analysis of the urban water-electricity demand nexus in the Midwestern United States}
\author[1]{Renee Obringer}
\author[2]{Rohini Kumar}
\author[1,3]{Roshanak Nateghi}

\affil[1]{Environmental and Ecological Engineering, Purdue University, West Lafayette, Indiana, USA}
\affil[2]{Department Computational Hydrosystems, Helmholtz Centre for Environmental Research - UFZ, Leipzig, Germany}
\affil[3]{School of Industrial Engineering, Purdue University, West Lafayette, Indiana, USA}

\date{}

\begin{document}

\maketitle

\begin{abstract}
Considering the interdependencies between water and electricity use is critical for ensuring conservation measures are successful in lowering the net water and electricity use in a city. This water-electricity demand nexus will become even more important as cities continue to grow, causing water and electricity utilities additional stress, especially given the likely impacts of future global climatic and socioeconomic changes. Here, we propose a modeling framework based in statistical learning theory for predicting the climate-sensitive portion of the coupled water-electricity demand nexus. The predictive models were built and tested on six Midwestern cities. The results showed that water use was better predicted than electricity use, indicating that water use is slightly more sensitive to climate than electricity use. Additionally, the results demonstrated the importance of the variability in the El Ni\~no/Southern Oscillation index, which explained the majority of the covariance in the water-electricity nexus. Our modeling results suggest that stronger El Ni\~nos lead to an overall increase in water and electricity use in these cities. The integrated modeling framework presented here can be used to characterize the climate-related sensitivity of the water-electricity demand nexus, accounting for the coupled water and electricity use rather than modeling them separately, as independent variables. 
\end{abstract}

\section{Introduction}
Urban areas are growing at an unprecedented rate \cite{TheWorldBank2010}, and as the cities grow, the water and electricity networks will become more stressed. This stress will be further exacerbated by climate change, which is expected to cause more variable conditions, including higher temperatures and increased frequency and intensity of drought events \cite{Dai2011}. Higher temperatures increase the energy demand for cooling which puts pressure on power plants as they have to generate more electricity to meet the increased demand. Higher energy generation will lead to more water withdrawals by the power plants \cite{Scanlon2013}. The higher temperatures and increased frequency of droughts will also put pressure on water resources. When water is scarce, power plants cannot withdraw as much as they need, and when water temperatures are higher than average they reduce the Carnot efficiency of thermal plants both of which stress the energy systems. Specifically, it has been shown that a 1$^\circ$C cooling water temperature difference can lower a plant's capacity by 0.15-0.5\% \cite{Cook2015a}. For instance, prolonged heat-waves and low water levels over the past few years forced a number of countries such as France, Spain, and Germany to limit power generation and occasionally shut-down power plants \cite{VanVliet2013}. In this sense, the electricity sector puts pressure on the water sector by requiring a large amount of withdrawals, and the water sector puts pressure on the electricity sector when there are shortages. This is compounded by climate change, which will put additional pressure on both sectors. This interconnectivity of the water and electricity sectors is known as the water-energy nexus. 

The water-energy nexus is a concept dating back to the 1990’s that was recently brought into the spotlight after the publication of several government studies and reports in the late 2000’s \cite{Hussey2012}. Since the release of these studies and reports there have been many initiatives surrounding the water-energy nexus, calling for researchers to evaluate the nexus and its impacts at various spatiotemporal scales and for numerous applications. The idea behind studying the nexus, as opposed to studying water and/or energy in isolation, is that the two systems are interrelated and there is a potential for missed information if one studies them separately.

There are a few different ways to study the water-energy nexus. Frequently, researchers look into the water that is used during electricity generation. An estimated 90\% of the electricity in the US comes from thermoelectric power plants, which require water for cooling \cite{Scanlon2013}. The amount of water withdrawn by these plants accounted for 40\% of the water withdrawals in the US during 2005 \cite{Kenny2009}, making these plants a crucial aspect to studying water availability in the US. For example, a study by Sovacool and Sovacool projected the electricity supply and demand across the US and found that 22 metropolitan regions will experience water shortages because of the increase in electricity generation \cite{Sovacool2009}. Globally, there is approximately 52 billion cubic meters of fresh water withdrawn every year solely for energy production \cite{Spang2014}. The main reason for these water shortages is the cooling technology used in the plants. A study by Macknick et al. showed that low-carbon energy options, like concentrated solar power or carbon sequestration coal plants, would still lead to increased water withdrawals, but options like photovoltaics and wind energy that do not require cooling, would lead to decreased water withdrawals \cite{Macknick2012}. These results were further demonstrated by a global study of the impacts of water-intensive renewable energy portfolios. Hadian and Madani showed that `greener' energy sources, such as hydropower and biofuels, require more water than traditional fossil fuel sources. In fact, depending on the renewable energy sources used in the future, the world could see a 37-66\% increase in water use for energy in the next few decades \cite{Hadian2013}. In addition to the water shortages that will likely be caused by increased power generation, thermoelectric or otherwise, there will be stress caused by climate change and increased drought frequency and intensity \cite{Dai2011}. A study by Scanlon et al. showed that droughts led to a 6\% increase in electricity demand, which led to a 9\% increase in water withdrawals \cite{Scanlon2013}. 

Another way to study the water-energy nexus is to assess the electricity it takes to treat and distribute water. In fact, it has been estimated that water-related energy use will increase in states that are already water stressed, such as Florida, Texas, and Arizona \cite{Sanders2012}. This will likely be caused by a shift towards energy-intensive technologies, such as desalination, that focus on increasing water supply in arid regions. However, extraction is not solely responsible for the electricity used in the water sector. A study performed by Tarroja et al. showed that it was critical to take the electricity needs of drinking water conveyance, treatment, and distribution, as well as wastewater treatment into account when determining greenhouse gas emissions for the state of California \cite{Tarroja2014a}. Within the distribution system, pumping accounts for 80\% of the energy consumption \cite{Sedlack2014}. This energy requirement is further exacerbated by leaks in the distribution system, which necessitate additional energy to maintain the flow through the system, ultimately putting stress on the electricity sector \cite{Derrible2017}. These results were mirrored at the city level as well. For example, in Tucson, Arizona, water has to be imported from the northern part of state. The energy needed to transport the water accounts for 36\% of the total, which is in addition to the 14\% of the total electricity consumption that is used to treat and distribute the water once it gets to Tucson \cite{Perrone2011}. This puts great strain on the electricity utilities, especially in the face of continued population growth. There were similar results from a review of global urban water-energy nexuses \cite{Lee2017}. This review showed that the accessibility of water resources was the main driver for increased energy intensity and greenhouse gas production because of the amount of energy required to reach less accessible water sources, such as groundwater or salt water \cite{Lee2017}. While the aforementioned studies demonstrate that it is important to focus on the water-energy nexus, many of them have focused on the supply side, not the demand side. 

Similar to the supply-side interdependencies discussed above, there are many aspects of water and electricity use that are interconnected. For example, watering landscapes, washing clothes, taking hot showers, and using a dishwater all require both water and electricity. These interdependecies, known as the residential water-electricity nexus, are critical for utilities trying to reduce peak load. Reducing peak load is desirable for both water and electricity utilities, as it lowers the likelihood of an outage and reduces operations and maintenance cost \cite{Hoque2014a}. However, it is uncommon for water and electricity utilities to work together to ensure a net reduction in both water and electricity use. Unfortunately, this lack of integration can lead to unforeseen consequences for one utility or the other. For example, in Phoenix, Arizona, residents were encouraged to plant xeric landscaping to reduce water use. However, this changed the microclimate such that people were using air conditioning more often, leading to an increase in electricity use \cite{Ruddell2014}. Similarly, a study in Brazil found that when households implemented rain barrels to reduce their dependence on the centralized water system, the overall electricity consumption increased by 4\% \cite{Vieira2016}. The authors explained that this increase in electricity consumption was likely due to the diseconomies of scale, and for rain barrels to be effective, they ought to be paired with other conservation measures to reduce the need for a centralized sewage treatment facility (e.g., greywater reclamation).  These studies, among others, demonstrate the need to study the water-electricity demand nexus in addition to studying water use and electricity use as two separate entities.

The purpose of this study was to offer a new modeling framework to simultaneously assess the climate sensitivity of the demand for water, electricity, and their nexus. Using the developed predictive model, we analyzed the climate sensitivity of the water-electricity demand nexus. Since the coupled water-electricity nexus model takes the interdependencies between the response variables into account, it was hypothesized that this multivariate modeling framework would predict the water and electricity use better than similar univariate models. To test this hypothesis, we applied our framework to six large-range cities in the Midwestern United States and evaluated the impacts of climate variability on the demand nexus. We anticipated, based on previous studies, that both local climatic variables, such as precipitation and temperature, and large climatic drivers, such as the El Ni\~no/Southern Oscillation index, would be important predictors of end-use demand for water and electricity. 

\section{Methods and Data}
\subsection{Algorithm Description}\label{s: method}
We used an advanced statistical approach of modeling the coupled or multivariate system to account for a complex interaction between water and energy demand nexus. Specifically, the predictive models of the residential energy and water demand were developed based on a multivariate extension of the gradient boosted regression trees algorithm \cite{Friedman2001}. Gradient boosted regression trees is an ensemble, tree-based method that takes advantage of the boosting meta-algorithm to increase the predictive accuracy \cite{Friedman2001}. The boosting meta-algorithm can be applied to a wide range of algorithms, including decision trees, and works by sequentially reducing the loss function and improving the bias of the base learners to yield a robust and accurate final prediction. Boosting is represented mathematically in the equation below.

\begin{equation}
    y(x)=\sum_i^{N}w_iC_i(x)
\end{equation}

\noindent Here $y(x)$ is the final model predicting the values of $x$, $N$ is the total number of iterations to be completed, $W_i$ is the weight of each prediction, and $C_i$ is the prediction of $x$ made at iteration $i$. 

Multivariate tree boosting, extends gradient boosted regression trees to a multivariate case, and enables the simultaneous estimation/prediction of multiple response variables \cite{Miller2016}. Specifcially, this algorithm iteratively builds trees that minimize the squared error loss for each response variable and maximize the covariance discrepancy in the multivariate response. In other words, at each iteration, a prediction is made and those observations that were poorly predicted are given more weight in the next iteration. Given the new weight of various observations, a new prediction is made and the weighting process repeated. This will continue until a certain number of iterations has been completed, at which point the weighted predictions will be combined as a final model. This algorithm has been used in a variety of predictive applications, ranging from psychological well-being \cite{Miller2016} to infrastructure resilience \cite{Nateghi2017}. A more detailed discussion on the algorithm can be found in Miller et al. \cite{Miller2016}

\begin{figure}[t]
\centering
\includegraphics[width=\linewidth]{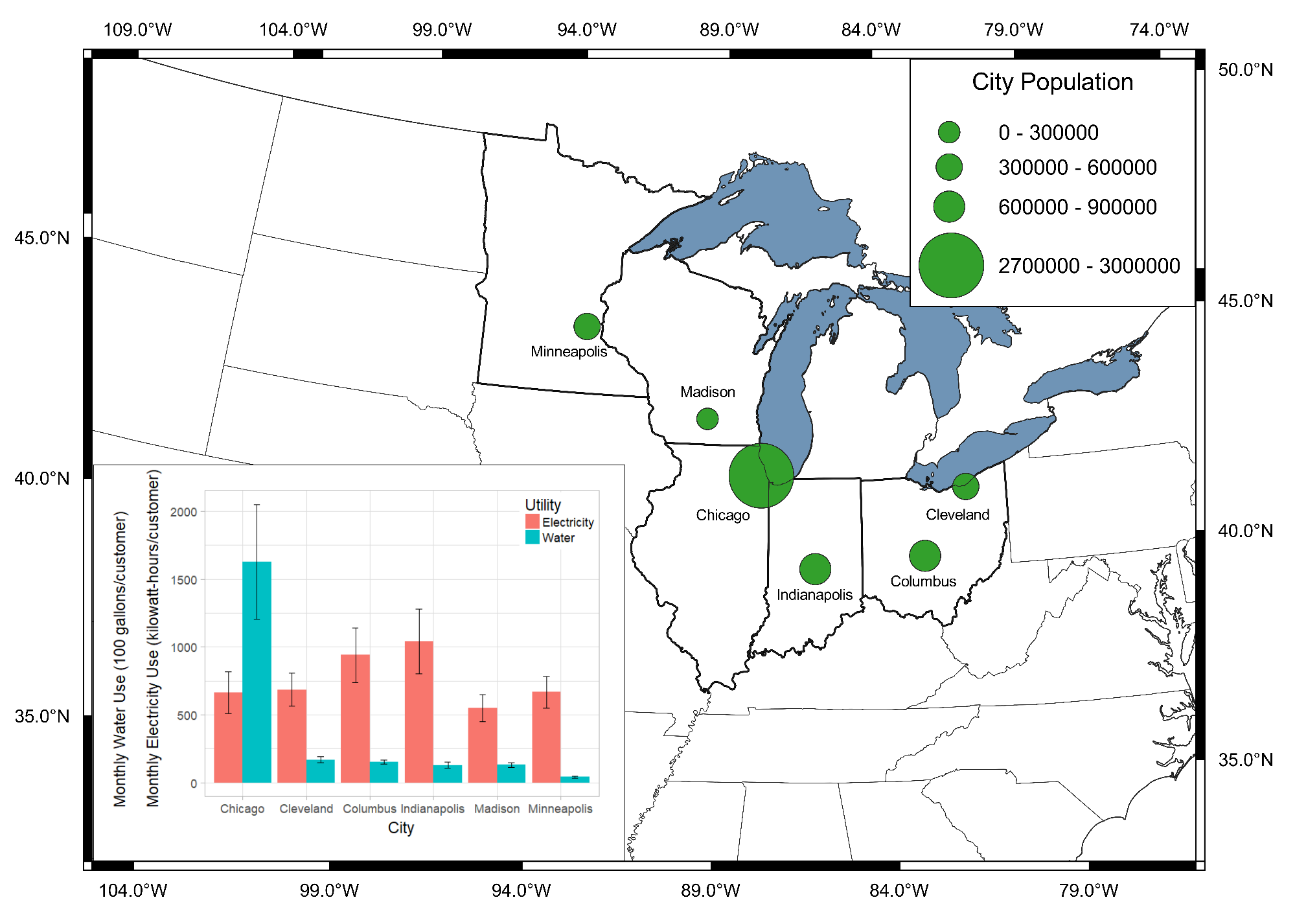}
\caption{Map of the cities chosen for this study. From left to right: Minneapolis (MN), Madison (WI), Chicago (IL), Indianapolis (IN), Columbus (OH), and Cleveland (OH). The inset plot shows the mean $\pm$ one standard deviation for both water and electricity use for each city. The water use is in 100 gallons/customer (or metered account) and the electricity use is in kilowatt-hours/customer.}
\label{fig1}
\end{figure}

\subsection{Site Description}
We tested our modeling framework in the Midwest region in the United States, a region which is a mix of agricultural and urban areas. In particular, there was a focus on the northern and eastern parts of the Midwest, including Ohio, Indiana, Illinois, Wisconsin, and Minnesota. Within this study area, depicted in Figure \ref{fig1}, six cities of varying population sizes were selected: Chicago (IL), Columbus (OH), Indianapolis (IN), Minneapolis (MN), Cleveland (OH), and Madison (WI). Chicago is located on the coast of Lake Michigan and has a population of approximately 2.7 million, making it the third largest city in the United States and the largest in the Midwest \cite{USCensusBureau2016a}. Columbus and Indianapolis, the second and third largest cities in the Midwest, have approximately 850 thousand people each \cite{USCensusBureau2016,USCensusBureau2016d}. Minneapolis is the most northern city selected for this analysis, with a population of about 413 thousand \cite{USCensusBureau2016c}. The remaining two cities, Cleveland and Madison, are the smallest of the selected study areas, with 385 thousand and 252 thousand people, respectively \cite{USCensusBureau2016e,USCensusBureau2016b}. Each city, though they have different demand patterns, will likely experience similar impacts of climate change, given their geographical proximity. In particular, it is likely that the Midwest region as a whole will have higher temperatures and more precipitation as $CO_2$ levels continue to rise \cite{IntergovernmentalPanelOnClimateChange2013}, which will in turn affect the urban water-energy demand nexus. 

\subsection{Data Description}
The data for this study was obtained from four major sources—the US Energy Information Administration (EIA), National Centers for Environmental Information (NCEI), National Oceanic and Atmospheric Administration (NOAA), and local sources. Specifically, monthly residential electricity use was obtained from the EIA, meteorological and climate data from the NCEI and NOAA, and residential water use from local sources. The climate data was collected from several meteorological towers stationed around each city and aggregated to get an average monthly value for each city between 2007 and 2016. Specifically, there were four active towers in Chicago, Columbus, and Minneapolis, three in Cleveland, and one in Indianapolis and Madison. Meteorological variables used in the analysis included temperature (dry bulb and dew point), relative humidity, wind speed, and precipitation. The El Ni\~no/Southern Oscillation index was also included in the analysis, as a large-scale climatic driver that has been shown to impact the climate of the Midwest \cite{Kluck2016}. A list of all the variables used in the model establishment can be found in Table \ref{table1} . 

\begin{table}[h]
\setlength{\tabcolsep}{15pt}
\renewcommand*{\arraystretch}{1.2}
\caption{The variables used in the coupled water-energy demand nexus model. The response variables are those that are being predicted based on the predictor variables.}
\label{table1}
\begin{tabular}{@{}lll}
	\bottomrule
	Variable Type & Variable Name & Units \\ [2px]
    \midrule
    \multirow{2}{4em}{Response} & Monthly Water Use & Gal/Customer \\ [2px]
    & Monthly Electricity Use & MWh/Customer \\ [2px]    
    \midrule
    \multirow{5}{4em}{Predictor} & Average Maximum Temperature & \(^\circ\)F\\ [2px]
    & Average Dew Point Temperature & \(^\circ\)F \\ [2px]
    & Average Maximum Relative Humidity & \%  \\ [2px]
    & Average Relative Humidity & \%  \\ [2px]
    & Average Maximum Wind Speed & mph \\ [2px]
    & Average Wind Speed & mph \\ [2px]
    & Monthly Accumulated Precipitation & in \\ [2px]
    & El Ni\~no/Southern Oscillaion index & -- \\ [2px]
    \bottomrule
\end{tabular}
\end{table}

In this study, there were two response variables: residential electricity use and residential water use, both normalized by the number of customers served, and 8 meteorological and climatic predictors. There was a focus on variables that are easily measured by meteorological stations because of the availability of such data, as well as the results of previous studies, which showed the importance of meteorological variables on water and electricity demand \cite{Balling2008, House-Peters2010, Nateghi2017, Mukhopadhyay2017a}. Similarly, it has been shown that the El Ni\~no/Southern Oscillation plays an important role in affecting hydroclimatic processes across the US, and reservoir levels in particular \cite{Obringer2018a}, and thus making it an important variable to include in the analysis of residential water use. 

\subsection{Statistical Modeling and Analysis}
To reduce the variance within the modeling framework, 5-fold cross validation was used so that at any given time, 80\% of the data was being used for training and 20\% was being held back for testing. This cross validation method was used in both tuning the model parameters as well as variable selection.

The performance of the model was assessed using two statistical measures: the out-of-sample root-mean-squared error (RMSE) and the out-of-sample coefficient of determination (\(R^2\)). RMSE provides a measure of error that heavily penalizes large deviations, making it ideal for prediction applications. The \(R^2\) value demonstrates the goodness of fit for the predictions made by the model. Both of these statistical measures were used to determine the out-of-sample error of the model. 

Additionally, a test for seasonality effects was performed on the electricity and water demand data. Specifically, in the seasonality analysis, the time series were decomposed and the seasonality components were subtracted from the original time series. This new time series was used as an input to the final model for an additional analysis. This analysis based on the de-trended data focused on demonstrating the effect of climate on the water-energy demand nexus independent of the natural seasonality present in the usage pattern. Finally, the results from the multivariate model were compared to results from a similar univariate model. Specifically, we used gradient tree boosting \cite{Friedman2001} to predict the water and electricity use as independent variables. Gradient tree boosting is the basis for multivariate tree boosting \cite{Miller2016}, thus the main difference between the multivariate and univariate algorithms is the consideration of response variable interdependencies. 

\section{Results}
\subsection{Model Performance and Variable Selection}\label{sec3.1}
To develop a predictive model of interdependent urban water and electricity demand, we leveraged the multivariate tree-boosting algorithm described in Section \ref{s: method}. In the initial training of model all the independent variables that could potentially affect water and/or electricity demand were included. Variable selection was then implemented by computing and ranking the independent variables based on their relative influence, which is done by calculating the percentage reduction in prediction error attributed to each variable. The final model included a reduced variable set---using the key predictors---such that the performance of the reduced model was statistically similar to that of the full model. 

The selected variables in the final model include maximum dry bulb temperature, average dew point temperature, average relative humidity, average wind speed, and the El Ni\~no/Southern Oscillation index. The selected variables are similar to previous studies on the sensitivity of water and electricity demand \cite{Balling2008, House-Peters2010, Mukhopadhyay2017a, Gotham, Mukhopadhyay}. The performance of the final model was assessed based on the coefficient of determination (\(R^2\)) and the root-mean-squared error (RMSE). These measures of error were calculated on the test set, or out-of-sample error. For each city there is an \(R^2\) and RMSE value for water and electricity demand. The results are summarized in Table \ref{table2}. 

\begin{table}[h!]
\setlength{\tabcolsep}{15pt}
\renewcommand*{\arraystretch}{1.5}
\caption{The model performance for each city for the final model run before the seasonality was removed from the response data.}
\label{table2}
\begin{tabular}{@{}lllll}
	\bottomrule
	\multirow{2}{3em}{City} & \multicolumn{2}{c}{Water Use} & \multicolumn{2}{c}{Electricity Use} \\ [2px]
    & \(R^2\) & RMSE & \(R^2\) & RMSE \\ [2px]
    \midrule
    Chicago & 0.47 & 0.731 & 0.76 & 0.499 \\ [2px]
    Columbus & 0.78 & 0.619 & 0.84 & 0.496 \\ [2px]
    Indianapolis & 0.83 & 0.491 & 0.87 & 0.385 \\ [2px]
    Minneapolis & 0.81 & 0.468 & 0.83 & 0.431 \\ [2px]
    Cleveland & 0.31 & 0.876 & 0.77 & 0.566\\ [2px]
    Madison & 0.71 & 0.623 & 0.77 & 0.450 \\ [2px]
    \bottomrule
\end{tabular}
\end{table}

\subsection{Seasonality Analysis}
\begin{figure}[h!]
\centering
\includegraphics[width=\linewidth]{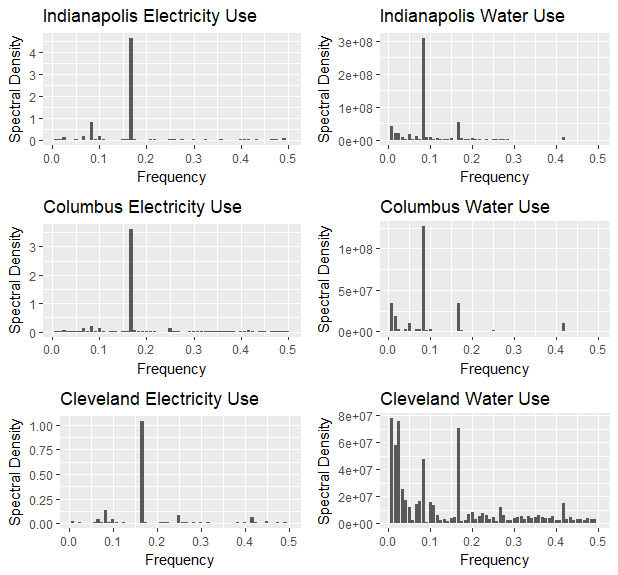}
\caption{Periodograms of a selection of the cities analyzed in this study. In these periodograms, the lone peaks demonstrate that seasonality is present at that frequency. Note that Cleveland has no apparent seasonality for water use.}
\label{fig2}
\end{figure}

The final step in the modeling process involved removing the seasonality from the response variables and rerunning the final model to determine the climate sensitivity of the `de-trended' residential electricity and water demand. When seasonality is present in the data, it often leads to better measures of accuracy and potentially biased predictive models. It is important to determine if there is a real impact due to climate or if seasonality in the response variables was responsible for higher predictive accuracies. The first part of this task was to test the extent to which seasonality was present in the data. Using the periodograms (see Figure \ref{fig2}), it was possible to confirm that there was a 6-month seasonality for electricity usage, which likely coincides with increased air conditioning use for the summer months in the Midwest. The water usage demonstrated a 12-month seasonality, indicating that there is also a significant (repeated) annual course of water use in the selected cities.

\noindent
 The exceptions to this observation were Chicago and Cleveland, which had no seasonality with respect to water use. Interestingly, Chicago and Cleveland also exhibited low sensitivity to climate with regard to water use in the final model run described in Section \ref{sec3.1}. This likely explains the low \(R^2\) values observed for these two cities, since seasonality usually increases the predictive ability of such models. After testing for seasonality within the electricity and water demand time series, the series were decomposed and the seasonality parts removed from the original data. This de-trended data was then used in the final model, as described above. Results from this study, shown in Table \ref{table3}, demonstrate that the seasonality was leading to higher predictive-ability, especially with respect to water demand. This can be seen in both \(R^2\) and RMSE values which are relatively inferior to the earlier results of predicting the overall water and energy demands (i.e., without discounting for the seasonal component). 

\begin{table}[h!]
\setlength{\tabcolsep}{15pt}
\renewcommand*{\arraystretch}{1.5}
\caption{The model performance for each city for the final model run after the seasonality was removed from the response data.}
\label{table3}
\begin{tabular}{@{}lllll}
	\bottomrule
	\multirow{2}{3em}{City} & \multicolumn{2}{c}{Water Use} & \multicolumn{2}{c}{Electricity Use} \\ [2px]
    & \(R^2\) & RMSE & \(R^2\) & RMSE \\ [2px]
    \midrule
    Chicago & 0.51 & 0.720 & 0.39 & 0.932 \\ [2px]
    Columbus & 0.62 & 0.894 & 0.31 & 0.975 \\ [2px]
    Indianapolis & 0.71 & 0.739 & 0.41 & 0.934 \\ [2px]
    Minneapolis & 0.55 & 0.761 & 0.42 & 1.113 \\ [2px]
    Cleveland & 0.23 & 0.910 & 0.34 & 0.943 \\ [2px]
    Madison & 0.34 & 0.925 & 0.30 & 1.003 \\ [2px]
    \bottomrule
\end{tabular}
\end{table}

\section{Discussion}
The goal of this study was to build a predictive model of the residential demand for water, energy and their nexus, and apply it in six Midwestern cities. Building off previous work in predicting water and electricity use, it was hypothesized that both response variables would be sensitive to climate variables. This was, for the most part, proved true throughout the analysis, although the level of sensitivity depended on the city in question. Additionally, our results established the higher climate sensitivity of the residential demand for water as compared to electricity.

Initially, it seemed as if electricity use was more sensitive to climate, since the out-of-sample $R^2$ values were higher for electricity use than water use in all the cities (Table \ref{table2}). This was especially true with the water use in Cleveland and Chicago. These two cities had a much lower \(R^2\) value for water use than all of the other cities, following the initial model run. This means that Cleveland's and Chicago's water use is less sensitive to climate than the other cities. However, additional analyses demonstrated that there was seasonality present in all of the cities, except the water use data for Cleveland and Chicago. 

Since seasonality often leads to an improved but biased prediction, it was important to re-run the final model with de-trended data. Following a test for seasonality using periodograms (see Figure \ref{fig2}), it was shown that Cleveland and Chicago had no seasonality for water use, while the remaining cities had 12-month seasonality. This likely explains the lower \(R^2\) for Cleveland's and Chicago's water use prediction, since the seasonality would have boosted the predictive accuracy and therefore the $R^2$ values. This was further confirmed when we reran the final model with data that had had the seasonality aspect removed. As shown in Table \ref{table3}, the non-seasonal model run led to lower \(R^2\) values for electricity use and water use, with the exception of Chicago's water use which had a slightly larger predictive accuracy after the seasonality was removed. This indicates that both water and electricity demand is less sensitive to climate than previously thought and the seasonality was in fact leading to a higher measure of accuracy. Interestingly, the results from this analysis show that following the decomposition and removal of the seasonality portion of the time series, the water use is better predicted than the electricity use. This demonstrates that water use is more sensitive to climate variables than electricity use. That being said, the goodness-of-fit is still somewhat low for most cities, which suggests that although climate has some impact on water use in the Midwest, it is likely that there are other more important factors, such as socioeconomic levels, or the presence of droughts and/or heatwaves, when predicting water demand. However, there is still a significant amount of variance that could be explained by the model, indicating that the final model is fairly accurate at predicting the climate-sensitive portions of the water and electricity demands. \\

\begin{figure}[h!]
\centering
\includegraphics[width=\linewidth]{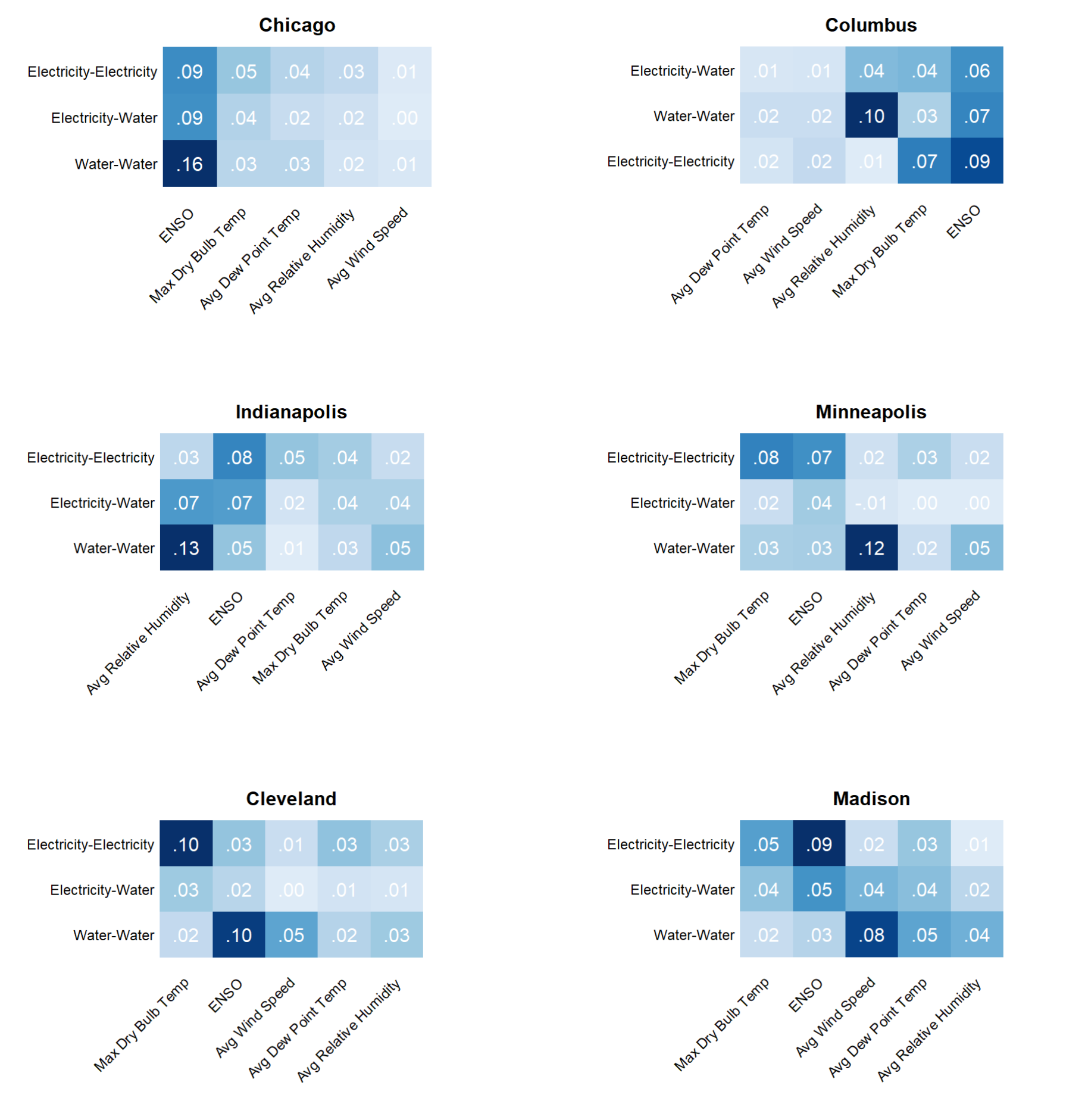}
\caption{Clustered heat maps showing the covariance explained by each predictor variable in each city, after the seasonality was removed from the dataset. The darker blues represent higher values of covariance explained, while the lighter blues represent less. The variables have been grouped using hierarchical clustering, a method used to group similar objects together. In this figure, predictors clustered together explain the covariance in similar outcome pairs. In this figure, the position of the variables on the axes is different for each city because each city has a different clustering outcome.}
\label{fig3}
\end{figure}

One of the advantages of multivariate tree boosting is the ability to determine the covariance explained by the predictors for each individual response variable and the nexus between response variables. This feature allows us to see what variables have the most impact on the water-electricity nexus and if those variables differ from those most greatly impacting water or electricity use alone. Figure \ref{fig3} shows the clustered heat maps of the covariance explained for each city. The results from the heat maps demonstrate the fact that the model itself is generalizable across different cities, but the results, including the important variables, will differ from city to city. For example, in the land-locked cities of Columbus, Indianapolis, and Minneapolis, average relative humidity explains the most covariance in water use. This is different than the coastal cities of Chicago and Cleveland, where the ENSO index explains much of the water use and relative humidity has less of an impact. This indicates that utility managers that are trying to reduce water use in Columbus or Indianapolis should focus on the days with high humidity, as that is when people are using the most water. Likewise, a manager in Chicago or Cleveland should focus their demand reduction efforts for strong El Ni\~no years. 

The impact of El Ni\~no---a large scale climate predictor---aligns with previous results focusing on the water supply side of urban water systems \cite{Obringer2018a}. In other words, the results from this study indicate that the El Ni\~no plays an important role in both aspects of urban water availability (supply and demand). The ENSO index also explains much of the covariance in the water-electricity nexus for each city. The El Ni\~no is a well-documented climate phenomenon that is also fairly easy to predict, making it an ideal variable to make more general or broad predictions. For example, a common ENSO-based prediction is the type of winter that a given region will have (e.g., a strong El Ni\~no usually leads to warmer, drier winters in the Midwest). Our modeling framework allows us to make a simple, first order forecast for the demand nexus based on large scale climate predictor. In other words, our results suggest that a strong El Ni\~no is more likely to lead to a higher water-electricity demand nexus. This knowledge would allow utility managers to prepare for the upcoming season based on the ENSO index that is determined on a monthly basis.

One of the goals of this work was to demonstrate the power of including both water and electricity use in the model as interdependent response variables. This was done through a model performance comparison of the multivariate tree boosting model and a univariate version: gradient tree boosting. The results from the univariate model run are shown in Table \ref{table4}. 

\begin{table}[h!]
\setlength{\tabcolsep}{15pt}
\renewcommand*{\arraystretch}{1.5}
\caption{The model performance ($R^2$) of the univariate model, gradient tree boosting, for each city after the seasonality was removed from the data.}
\label{table4}
\begin{tabular}{@{}lcc}
	\bottomrule
	City & Water Use & Electricity Use \\ [2px]
    \midrule
    Chicago & 0.50 & 0.32 \\ [2px]
    Columbus & 0.53 & 0.26 \\ [2px]
    Indianapolis & 0.64 & 0.29 \\ [2px]
    Minneapolis & 0.55 & 0.32 \\ [2px]
    Cleveland & 0.36 & 0.28 \\ [2px]
    Madison & 0.37 & 0.28 \\ [2px]
    \bottomrule
\end{tabular}
\end{table}

Our results show that both algorithms resulted in similar patterns between the water and electricity use predictive accuracies. For the most part, the water use was predicted more accurately than the electricity use, indicating that it is slightly more climate sensitive than electricity use. Additionally, the relative performance of the various cities matched between the univariate and multivariate models. For example, in both algorithms, Indianapolis's water use was predicted the most accurately, while Cleveland's was the least accurately predicted. Overall, however, the multivariate model was better at predicting the two demands than the univariate model, with the exception of Cleveland's and Madison's water use. The main difference between the univariate and multivariate models was the inclusion of response variable interdependencies within the multivariate model. This is indicative that, in most cases, the consideration of the interconnectivity between water and electricity use improves the final prediction of both water and electricity use. Of the cities tested as a part of this analysis, Cleveland's and Madison's water demands were more accurately predicted by the univariate model, which suggests loose coupling between water and electricity use in those cities than the other cities studied. Additional research is necessary to determine the reason behind this reduced coupling between the energy and water demands, however, it is possible that the population size is playing a role, since Cleveland and Madison are the smallest cities included in this study.

\section{Conclusions}
The purpose of this study was to build a model that identified the key predictors of the climate-sensitive portion the urban residential demand for water, electricity and their nexus, using the multivariate tree boosting algorithm. In this study, there were two response variables: water use and electricity use, and initially eight predictor variables, which was reduced to the five most important variables after the first model run. The model was tested on six Midwestern cities of variable size, demonstrating the generalizability of the model. The results of the study showed that the model performed well in predicting the climate-sensitive portion of the water and electricity demand, although it predicted the water use better than the electricity use. It is likely that there are other variables that are important for predicting water and electricity use, such as socio-demographic data, which were not included in this study. The results indicated that water and electricity use are sensitive to climate variables, and will likely be affected by future climate change. The impact of the El Ni\~no/Southern Oscillation was especially important in each city for explaining the covariance in the water-electricity nexus, and even in the individual response variables. These results can be used by utility managers that are interested in tailoring conservation interventions to the times at which they will be most effective. For example, focusing on conservation during a strong El Ni\~no will likely be more effective and result in greater reductions than the same campaign during a weak El Ni\~no. Finally, we compared the model performance to a similar univariate algorithm, known as gradient tree boosting. Our results demonstrated that in the majority of cities studied, multivariate tree boosting outperforms the univariate version. Since the main difference between the algorithms is the inclusion of multiple interdependent response variables, we recommend that future studies, especially in the Midwest, focus on modeling the water-electricity nexus, even if they are only interested in one or the other. 

\bibliographystyle{ieeetr}
\bibliography{demandNexus}

\end{document}